\pdfoutput=1
\documentclass[11pt]{article}
\usepackage[table]{xcolor}
\usepackage{tcolorbox}
\usepackage{acl}

\usepackage{times}
\usepackage{latexsym}

\usepackage{booktabs}
\usepackage{multirow}

\usepackage[T1]{fontenc}
\usepackage[utf8]{inputenc}

\usepackage{inconsolata}
\usepackage{textgreek}
\usepackage{enumitem}
\usepackage{graphicx} 
\usepackage{svg}
\usepackage{fancybox}

\usepackage{lipsum}

\newcommand{\myknowledgestore}{\textsc{MediaRef}}

\newcommand{\best}{\cellcolor{blue!30}}
\newcommand{\secondbest}{\cellcolor{blue!10}}

\title{\myknowledgestore{}: A Public Knowledge Store for Media Background Checks}

\author{
Benjamin Nichols\thanks{Work done while at Cardiff University},
Michael Schlichtkrull$^\diamond$,
Nedjma Ousidhoum$^\triangle$\\
$^\triangle$ Cardiff University, $^\diamond$ Queen Mary University of London\\
Correspondence: \texttt{OusidhoumN@cardiff.ac.uk}
}

\begin{document}
\maketitle
\begin{abstract}
LLM-based retrieval-augmented generation (RAG) is increasingly used for automated fact-checking (AFC) and related tasks. By grounding LLM outputs in retrieved evidence, RAG-based systems provide transparent justifications while allowing external information to be updated independently of the underlying model. However, existing approaches often assume retrieved evidence is reliable, although real-world information may be conflicting, outdated, and can originate from unreliable or biased sources.
Recent work on \textit{source-critical reasoning} addresses this challenge through media background checks (MBCs) \cite{Schlichtkrull-2024:MBCs}, which assess the credibility of evidence sources to support downstream fact verification. However, generating MBCs relies on costly proprietary search APIs, limiting reproducibility.

To mitigate this issue, we introduce \myknowledgestore{},\footnote{\myknowledgestore{} will be publicly available upon acceptance.}, a publicly available knowledge store of web-sourced documents that enables reproducible, low-cost evaluation of MBC generation across 200 media sources.
We describe a reproducible methodology for constructing and updating \myknowledgestore{}, assess widely used LLMs on the MBC generation task, and demonstrate that our knowledge store supports higher-quality MBC generation through both automatic and qualitative evaluation.

\end{abstract}

\section{Introduction}
\label{sec:Introduction}
\begin{figure*}
    \centering
    \includegraphics[width=\linewidth]{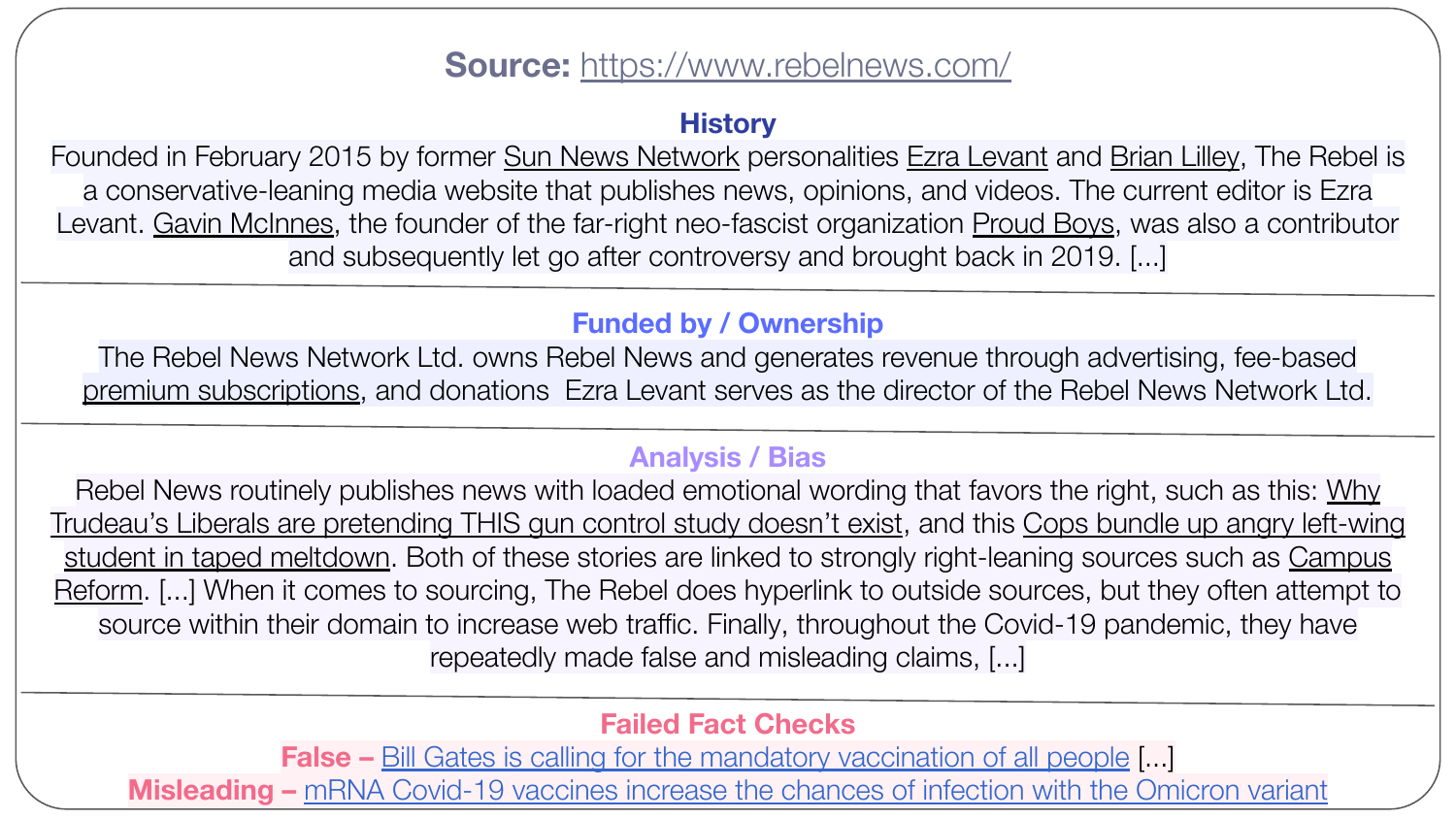}
    \caption{
An example of a media background check (MBC) for the Rebel News outlet, adapted from Media Bias/Fact Check. The full text is available at \url{https://mediabiasfactcheck.com/the-rebel/}.
    }
    \label{fig:mbc_example}
\end{figure*}

Scrutinising evidence sources is central to assessing the reliability of claims and identifying issues such as misleading information, omitted context, or unsupported statements, particularly in high-stakes domains such as journalism, public health, and policymaking \citep{GravesAmazeen-2019, Warren-2025:Fact-Checkers_Requirements}. However, manually evaluating the evidence underlying a claim is time-consuming and resource-intensive, especially given the growing volume of online information \citep{Allcot-2017:SocialMisinfo}. For instance, verifying a single claim can require a full day of work for a journalist, while more complex investigations may take several days \citep{Hassan-2015}. Consequently, automated fact-checking (AFC) has emerged as an active area of research \citep{Guo-Survey:2022}, with approaches ranging from verification over fixed corpora to open-domain systems that dynamically retrieve evidence from the web.

Recent advances in LLM-based retrieval-augmented generation (RAG) systems have further broadened the scope of AFC \citep{Lewis-2020:RAG, Schlichtkrull-2024:AVeriTeC-Task}. However, research on the trustworthiness of LLM-based systems has largely focused on issues such as prompt security, adversarial attacks, and hallucinations \citep{Vassilev-2024:AdversarialML, Lent-2025:NLPSec}. Comparatively less attention has been paid to the reliability of the external information sources on which retrieval-based systems depend. That is, retrieved evidence is often treated at face value, despite the possibility that sources may be conflicting, outdated, incomplete, or intentionally deceptive \citep{Hong-2024:GullibleLLMs, Ge-2025:MBCsEval}. This raises the risk of \textit{attacks-by-content} \citep{Schlichtkrull-2025:AttacksByContent}, whereby false or manipulative information propagates through retrieval pipelines and influences downstream model outputs.
One promising direction for explicitly considering source credibility is the use of \textit{media background checks} (MBCs) \citep{Schlichtkrull-2024:MBCs}. MBCs are summaries of aspects that may affect perceptions of a source's trustworthiness, including political orientation, factual reliability, editorial practices, and ownership (e.g. \autoref{fig:mbc_example}). By providing contextual information about evidence sources, MBCs can help both humans and automated systems make more informed judgements about retrieved information. However, existing approaches to MBC generation typically rely on proprietary search engine APIs to retrieve source information at inference time, introducing financial costs, limiting reproducibility, and creating variability as search engine rankings and retrieval systems evolve.

To overcome these limitations, we introduce \myknowledgestore{}, a collection of web-sourced documents covering 200 news sources that supports MBC generation. By decoupling information retrieval from MBC generation, \myknowledgestore{} provides a reproducible, low-cost alternative to proprietary search APIs while reducing variability arising from changes in retrieval behaviour over time. We describe a systematic methodology for constructing the resource, enabling the collection to be kept up to date as web content evolves.
We use \myknowledgestore{} to evaluate the MBC-generation capabilities of several LLMs and examine the impact of additional source-related evidence on generation quality. We find that access to \myknowledgestore{} produces MBCs that contain more accurate information about news sources without increasing the incidence of misleading content. We further develop a qualitative evaluation framework for MBCs based on four criteria---\emph{clarity}, \emph{relevance}, \emph{informativeness}, and \emph{verifiability}---motivated by users' need to assess the trustworthiness of MBCs. This analysis complements prior work on the usefulness of MBCs \citep{Schlichtkrull-2024:MBCs} by identifying, at a finer level of granularity, which characteristics contribute to their utility and when MBCs may fall short across these dimensions.

\section{Related Work}

AFC is typically divided into three stages: \emph{claim detection}, which identifies claims to fact-check; \emph{evidence retrieval}, which gathers information to support or refute a claim; and \emph{claim verification}, which determines the overall truthfulness of a claim \citep{Guo-Survey:2022}. Earlier AFC datasets \citep{Thorne-2018:Fever-Data, Jiang-2020:HoVer} relied on evidence from a single closed source, typically Wikipedia, assuming the source to be trustworthy. However, this risks propagating biases present in the underlying source \citep{Baly-2018:NewsSourceBias}. More recent work instead retrieves online evidence dynamically during model runtime using search APIs \citep{Schlichtkrull-2023:AVeriTeC-Data}. Further, latest systems use retrieval-augmented generation (RAG) to retrieve evidence and generate human-understandable \emph{justifications} that explain how the retrieved evidence supports a verdict \citep{Schlichtkrull-2024:AVeriTeC-Task}. Professional fact-checkers report that such justifications enable audiences to scrutinise the reasoning process, thereby promoting trust \citep{Warren-2025:Fact-Checkers_Requirements}.

Despite these advances, dynamically retrieving online evidence remains constrained by the financial cost of search APIs \citep{Schlichtkrull-2024:AVeriTeC-Task}, potentially limiting both future research and practical adoption by resource-constrained fact-checking organisations \citep{Poynter-2026:State_Of_Fact_Checking}. To reduce this barrier, shared tasks such as AVeriTeC \citep{Schlichtkrull-2024:AVeriTeC-Task}, AVeriTeC~2.0 \citep{Akhtar-2025:AVeriTeC2.0}, and AVerImaTeC \citep{Cao-2026:AVerImaTeC} have provided participants with stores of pre-retrieved online evidence, avoiding the need for costly API calls. Participating teams have reported additional benefits, including improved reproducibility and robustness against the unexpected disappearance of online information over time \citep{Rothermel-2024:TUDA_MAI}. However, retrieved evidence is often treated as \textit{credible} based on proxy signals such as search ranking position, despite no guarantee that highly ranked sources are free from misinformation. If left unresolved, conflicting evidence can substantially degrade the performance of RAG systems \citep{Hong-2024:GullibleLLMs, Ge-2025:MBCsEval}.
To address this issue, \citet{Schlichtkrull-2024:MBCs} introduced \emph{Media Background Checks} (MBCs) for AFC. They proposed a two-stage RAG pipeline in which evidence describing characteristics of a source associated with (in)credibility is first retrieved and then used by an LLM to generate an MBC. Building on this work, \citet{Ge-2025:MBCsEval} explored how MBCs could be integrated into AFC pipelines and examined their impact on the fact-checking process through qualitative analysis. In this paper, we build on this line of work by introducing \myknowledgestore{}, a curated knowledge store for media background checks that enables reproducible evidence access across multiple LLMs without relying on live search APIs. We further propose a qualitative analysis that goes beyond assessing the utility of MBCs alone \cite{Schlichtkrull-2024:MBCs}, and instead examines fine-grained characteristics that affect their perceived usefulness.

\section{\myknowledgestore{} Creation}
\subsection{Media Background Checks}
Media background checks (MBCs) (e.g., \autoref{fig:mbc_example}) are textual descriptions of media outlet characteristics that may affect the credibility of their reporting. These characteristics include, for example, an outlet's history, ownership, political bias, and prior fact-checking record. MBCs are supported by URLs linking to external evidence sources, such as relevant news articles or other MBCs, which readers can consult to assess their reliability. Beyond supporting human readers, MBCs provide a signal of source credibility that can be incorporated into downstream systems, such as automated fact-checking models that rely on web-sourced information, helping them prioritise trustworthy evidence over misinformation.
\subsection{Dataset Source}

We use Media Bias/Fact Check (MB/FC)\footnote{\url{https://mediabiasfactcheck.com}} as a source of gold-standard MBCs. MB/FC is an independent US-based organisation that aims to equip news consumers to identify and defend against misinformation by \textit{``understanding the bias and credibility of the sources they consume''}. Its MBCs are written by volunteers following a published methodology\footnote{\url{https://mediabiasfactcheck.com/methodology/}} and cover outlets ranging from small local newspapers to international organisations.
Although MB/FC is frequently used as a ground-truth source in studies of source credibility \citep{Baly-2019:MB/FC, Nakov-2024:MB/FC-Survey}, prior work typically incorporates the credibility and political bias ratings provided by MB/FC rather than the MBC text itself. We therefore use the MB/FC dataset released by \citet{Schlichtkrull-2024:MBCs}, from which we randomly sample 200 outlets from the test split.

\begin{figure}[t]
    \centering
    \includegraphics[width=\linewidth]{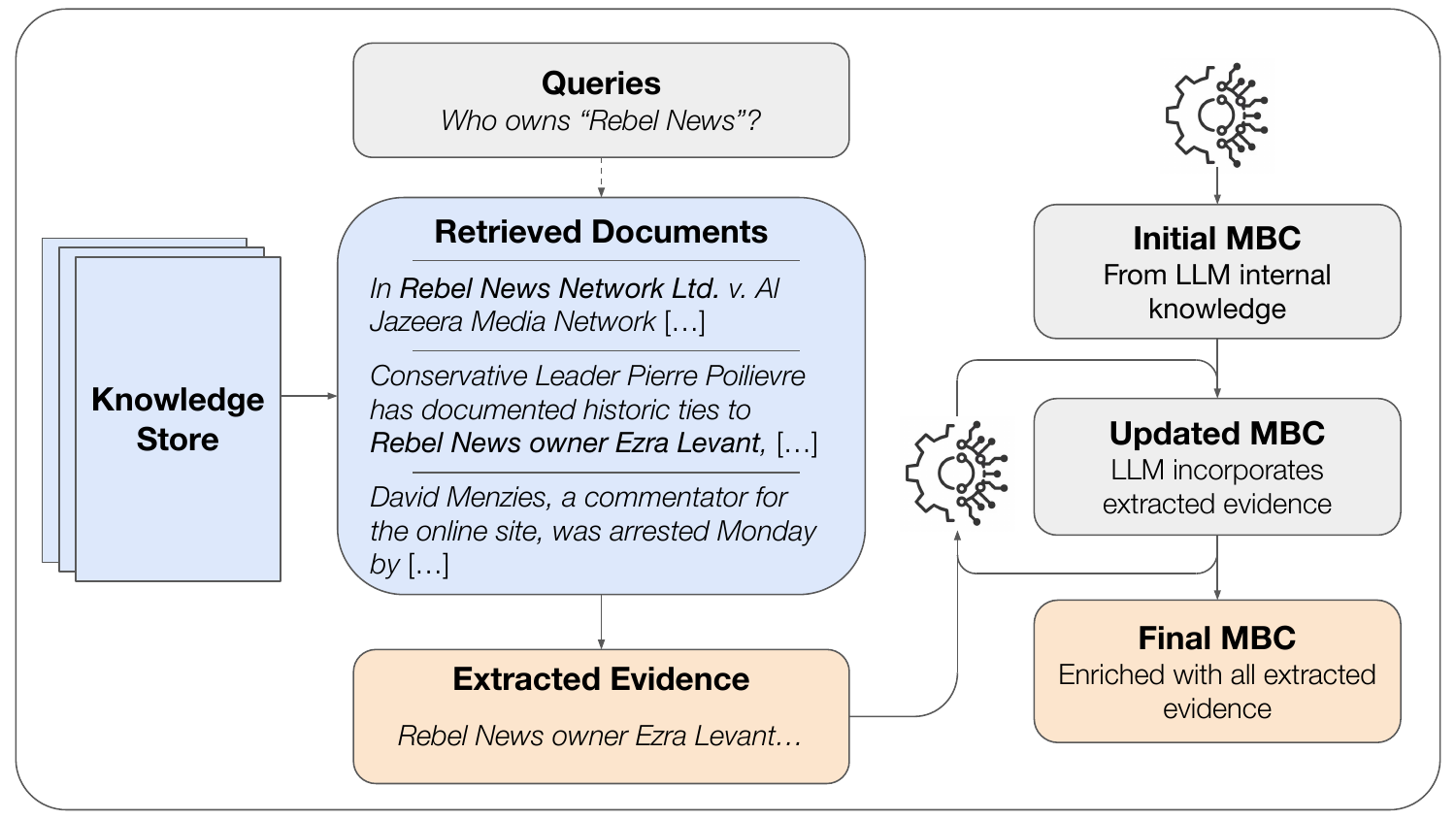}
    \caption{
    \textbf{MBC generation method for a given news source using external information.} We use targeted queries to retrieve relevant documents, and a question-answering model extracts passages of supporting evidence. An LLM generates an initial background check for the news source and updates it by incorporating the supporting evidence found.
    }
    \label{fig:mbc_generation}
\end{figure}

\subsection{Online Evidence Retrieval}

To gather background information for each news outlet required to generate an MBC, we construct a set of search queries targeting key aspects of a news outlet's background, including funding, ownership, bias, endorsements, and general descriptions. The queries follow the pattern \texttt{``source'' news <topic>}, where \texttt{source} is replaced by the outlet name and \texttt{<topic>} is one of {\texttt{funding}, \texttt{about}, \texttt{ownership}, \texttt{endorsement}, \texttt{bias}}. To minimise off-topic results, all queries include the keyword \texttt{news} and require an exact match to the outlet name. For each outlet, we then submit all queries to the Google Search API using default parameters and collect the top 10 results per query. Our retrieval process can be replicated to obtain more recent sources.

\paragraph{Quality Control}

To ensure sufficient coverage for reproducing the gold MBCs provided by \citet{Schlichtkrull-2024:MBCs}, we remove duplicate URLs and additionally include all external URLs cited in the gold MBC texts, as in \citet{Schlichtkrull-2024:AVeriTeC-Task}, merging them with the retrieved URLs. Furthermore, to reduce the risk of data leakage, we remove URLs pointing to webpages that are likely to reference the gold MBCs. We exclude all URLs matching a blacklist of websites known to cite MB/FC content, following \citet{Schlichtkrull-2024:MBCs}.

We then scrape the remaining webpages using the open-source web-scraping tool \texttt{trafilatura} \cite{Barbaresi-2021:Trafilatura}. This results in 21,921 non-empty documents, which form the final knowledge store.

\section{Experimental Setup}

We investigate the impact of using \myknowledgestore{} as a source of evidence for LLM-generated MBCs. We experiment with models from OpenAI \citep{Radford-2019:GPT3, OpenAI-2024:GPT4-Report, Singh-2025:GPT-5-System-Card}, Qwen \citep{Yang-2024:Qwen-2, 2025:Qwen-3}, Llama \citep{Meta-2024:Llama-3}, Mistral \citep{Jiang-2023:Mistral7b}, and Anthropic \citep{Anthropic-2024:Claude-3.5}. This selection includes both state-of-the-art proprietary systems and smaller open-source alternatives that are more accessible to researchers and fact-checking practitioners, allowing us to assess current LLM capabilities on the MBC generation task.

\subsection{Models}

We compare MBCs generated with and without information retrieval and evaluate the resulting differences in quality using two prompting strategies described below. The full prompts are provided in \autoref{appendix:Prompts}.

\paragraph{Prompting Without Information Retrieval}
An MBC is generated in a single LLM call without access to external information. Given the name of a news outlet, the model is prompted in a zero-shot setting to generate a brief, itemised summary of information associated with the outlet.

\paragraph{Prompting With Information Retrieval (+IR)}
First, an initial MBC is generated using the zero-shot prompting strategy described above. The MBC is then refined using evidence retrieved from \myknowledgestore{} through the following process.

\textbf{(1) Retrieval.} Relevant documents are retrieved using BM25 \citep{Robertson-1994:BM25Okapi}. We define six keyword-based queries targeting recurring topics in MBCs, such as outlet history and funding (see \autoref{appendix:SearchQueries}), and retrieve the top 30 highest-scoring documents for each query.

\textbf{(2) Evidence Extraction.} The retrieved documents are filtered to identify passages likely to contain credibility-relevant information. Each keyword query is reformulated as a question; for example, \emph{``source name'' funding} becomes \emph{How is ``source name'' funded?}. A DeBERTa-based question-answering model \citep{He-2021:DeBERTa} is then used to extract answers from the retrieved documents. If an answer is found, the corresponding text span is retained as evidence.

\textbf{(3) MBC Update.} The extracted evidence is used to update the initial MBC. Document snippets are processed individually, and the MBC is revised iteratively after each snippet to reduce the risk of exceeding the model's context window. The model is instructed not to remove existing points, thereby minimising information loss.

\subsection{Evaluation Metrics}
To compare generated and gold-standard MBCs \citep{Schlichtkrull-2024:MBCs}, we use the following metrics:
\begin{itemize}[noitemsep,nolistsep]
    \item \textbf{ROUGE-L} which evaluates the longest common subsequence of words between the generated and reference texts \citep{Lin-2004:ROUGE}.
    
    \item \textbf{METEOR} which uses stemming and synonym matching to evaluate word-level overlap \citep{Banerjee-Lavie-2005:METEOR}.
    
    \item \textbf{Fact Recall} and \textbf{Error Rate} \citep{Schlichtkrull-2024:MBCs} based on the FActScore framework \citep{Min-2023:FActScore}. A GPT-3.5-Turbo model decomposes a gold-standard MBC into \emph{atomic facts} using a gap-filling template (e.g. \emph{The usual audience of ``source name'' is \_\_}; see \autoref{appendix:Atomic Facts}). It then verifies whether each extracted fact is entailed, contradicted, or neither by both the generated and gold-standard MBC texts.

    \begin{itemize}[noitemsep,nolistsep]
        \item \textbf{Fact Recall} is the proportion of atomic facts for which the generated MBC and the gold-standard MBC \textbf{\textit{agree}} on entailment.
        \item \textbf{Error Rate} is the proportion of atomic facts for which the generated MBC and the gold-standard MBC \textbf{\textit{disagree}} on entailment.
    \end{itemize}
\end{itemize}

\section{Results}
\begin{table*}[ht]
\begin{tabular}{@{}lccccc@{}}
\toprule
\centering
\texttt{\textbf{Models}} & \texttt{\textbf{Fact Recall}} & \texttt{\textbf{Error Rate}} &
\multicolumn{1}{l}{\texttt{\textbf{METEOR}}} &
\multicolumn{1}{l}{\texttt{\textbf{ROUGE-L}}} &
\multicolumn{1}{l}{\texttt{\textbf{\#Tokens}}} \\
\midrule
\texttt{gpt-3.5-turbo-0125} & 26.86\% & 8.21\% & 12.06\% & 13.66\% & 124.1 \\
\texttt{gpt-3.5-turbo-0125 + IR} & 28.24\% & 8.13\% & 15.95\% & 14.21\% & 200.5 \\
\texttt{gpt-4o-mini} & 28.33\% & 8.29\% & 15.82\% & 12.37\% & 248.5 \\
\texttt{gpt-4o-mini + IR} & \best\textbf{29.85\%} & 8.96\% & 18.36\% & 12.61\% & 342.4 \\
\texttt{gpt-5-mini} & 24.78\% & \best\textbf{4.80}\% & 18.20\% & 11.31\% & 324.2 \\
\texttt{gpt-5-mini + IR} & 27.83\% & \secondbest\textbf{5.48\%} & \best\textbf{23.89}\% & 11.35\% & 683.3 \\
\texttt{qwen-2.5-72b-instruct} & 24.45\% & 6.90\% & 10.79\% & 9.94\% & 175.5 \\
\texttt{qwen-2.5-72b-instruct + IR} & 24.62\% & 7.58\% & 16.82\% & 11.09\% & 393.3 \\
\texttt{qwen-3-32b-instruct} & 24.91\% & 9.78\% & 16.71\% & 11.22\% & 294.1 \\
\texttt{qwen-3-32b-instruct + IR} & 26.22\% & 9.79\% & 19.35\% & 11.12\% & 441.3 \\
\texttt{llama-3.3-70b-instruct} & 28.98\% & 9.66\% & 9.67\% & 10.57\% & 115.2 \\
\texttt{llama-3.3-70b-instruct + IR} & \secondbest\textbf{29.03\%} & 9.76\% & 16.90\% & 12.11\% & 321.5 \\
\texttt{mistral-7b-instruct-v0.3} & 28.07\% & 10.54\% & 22.28\% & \best\textbf{15.57\%} & 383.7 \\
\texttt{mistral-7b-instruct-v0.3 + IR} & 27.49\% & 10.44\% & \secondbest\textbf{22.69\%} & \secondbest\textbf{14.67\%} & 480.8 \\
\texttt{claude-3.5-haiku} & 26.66\% & 7.32\% & 10.53\% & 9.95\% & 160.2 \\
\texttt{claude-3.5-haiku + IR} & 26.95\% & 7.15\% & 12.30\% & 10.14\% & 214.6 \\ \bottomrule
\end{tabular}
\caption{
\textbf{Results for MBC generation using LLMs.}
Metrics are reported for MBCs generated both without and with information retrieval (IR) using \myknowledgestore{}, along with the average token count of generated MBCs (\texttt{\#Tokens}) per model.
For error rate, lower values are better; for all other metrics, higher values are better.
Best scores are shown in blue, and second-best scores in light blue. Overall, MBCs generated with IR outperform those without IR across all metrics except error rate, where performance remains similar or degrades slightly.}
\label{table:Results}
\end{table*}
\subsection{Automatic Evaluation}
We report the experimental results in \autoref{table:Results}. Using \texttt{gpt-3.5-turbo}, our system achieves fact recall and error rate scores comparable to the baseline results reported by \citet{Schlichtkrull-2024:MBCs} on the same model (28.2\% vs.\ 26.1\% fact recall and 8.1\% vs.\ 6.3\% error rate, respectively). This suggests that using \myknowledgestore{} provides a reliable proxy for model performance on recent evidence. Among the evaluated models, \texttt{gpt-4o-mini} achieves the highest fact recall score, followed by \texttt{llama-3.3}.

Overall, the results suggest that MBC generation remains challenging, with relatively low fact recall across all models. Notably, state-of-the-art systems such as \texttt{gpt-5-mini} do not substantially outperform smaller open-source models, including \texttt{llama-3.3} and \texttt{mistral-7b}, in accurately reflecting facts about news sources. Error rates are also broadly similar, although \texttt{gpt-5-mini} exhibits a noticeably lower error rate than the other OpenAI models. According to \citet{Singh-2025:GPT-5-System-Card}, \texttt{GPT-5} was trained to reduce factual errors in RAG settings, for example by minimising hallucinations, which may explain this improvement. Importantly, higher fact recall does not necessarily correspond to lower error rates, highlighting the need to evaluate both the amount of correct and incorrect information generated. That is, focusing exclusively on factual content risks overlooking errors that may subsequently distort credibility judgements about news sources.

Introducing information retrieval generally improves fact recall, indicating that evidence retrieved from the knowledge store helps extend a model's knowledge of a news source. However, the magnitude of this gain, as well as changes in token count, vary substantially across LLMs. Since evidence retrieval is independent of the choice of the LLM, these differences suggest that models differ in how effectively they incorporate retrieved information into their MBCs. We qualitatively investigate the potential causes of this variation in the next section.

\subsection{Human Evaluation of MBC Generation Quality}
\label{sec:Qualitative_Eval}
We complement our automatic evaluation with a human evaluation to better understand why models may struggle to reflect the information in the gold-standard references. In addition, since MBCs are intended to support source criticism, we assess whether human readers perceive them as providing useful and verifiable information.

We define four criteria for evaluation: \emph{clarity}, \emph{relevance}, \emph{informativeness}, and \emph{verifiability}. The first three are adapted from the criteria for fact-checking questions proposed by \citet{Ousidhoum-2022:Varifocal}. We introduce \emph{verifiability} to capture whether an MBC helps users assess credibility by pointing to additional trustworthy sources \citep{Warren-2025:Fact-Checkers_Requirements}.
All criteria are rated on a Likert scale from 0--3, except for clarity, which is rated from 0--2. A full description of the annotation scheme is provided in \autoref{appendix:Annotation Criteria}.
\paragraph{Clarity}
Individual points should be comprehensible both on their own and in the context of the full text, and the MBC should not contain inconsistencies or self-contradictions.
\paragraph{Relevance}
All facts mentioned in the MBC should relate to the target news source or closely connected entities and topics.
\paragraph{Informativeness}
The MBC should provide information that meaningfully influences a reader's perception of a source's overall credibility.
\emph{Highly informative} background checks directly indicate the source's credibility and political bias (or absence of bias), and justify why.
\emph{Weakly informative} checks only describe characteristics loosely relevant to source credibility, such as general background information, with unclear implications for credibility.
\paragraph{Verifiability}
Sufficient evidence should be provided to support factual claims, or enough detail should be included so that one can reasonably locate the evidence independently.
This should be determined based solely on judgements of how easily the facts could plausibly be verified (i.e., without the use of a search engine or an LLM).

\begin{figure*}[t]
    \centering
    \includegraphics[width=\textwidth]{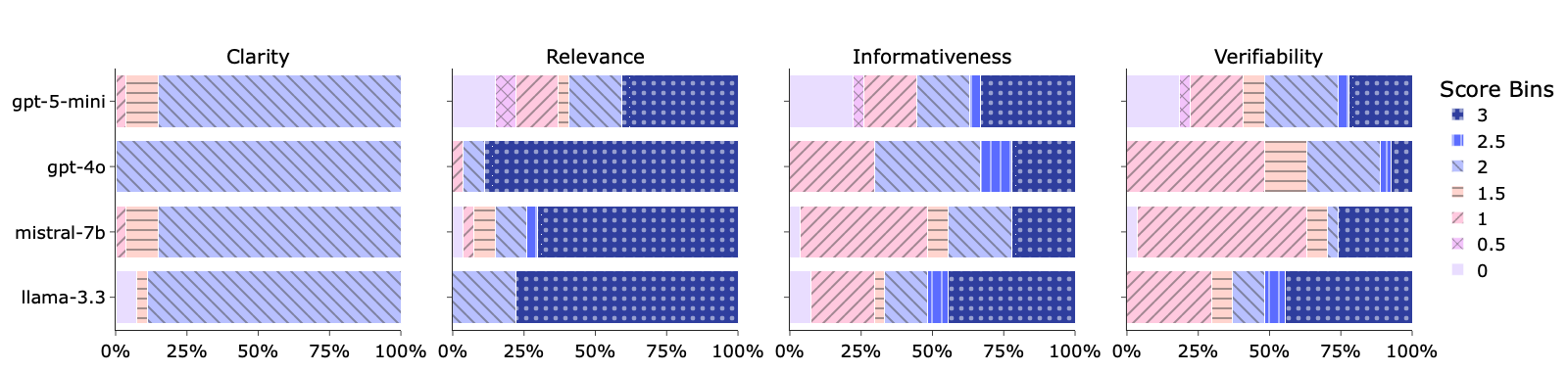}
    \caption{Qualitative analysis results for Clarity, Relevance, Informativeness, and Verifiability. Colours encode score magnitude (low → high, i.e., 0 → 3), consistently across all models. Note that the maximum Clarity score is 2, and the values shown are the average scores assigned by our two expert annotators.}
    \vspace{-0.2cm}
    \label{fig:qualitative_analysis}
\end{figure*}
\subsection{Human Evaluation Setup}

We randomly sample 108 LLM-generated MBCs (27 per model) from the \texttt{gpt-4o-mini}, \texttt{gpt-5-mini}, \texttt{llama-3.3-70b-instruct}, and \texttt{mistral-7b-instruct} IR-augmented models (i.e., +IR in \autoref{table:Results}) to investigate how information retrieved from the knowledge store affects the quality of the generated MBCs. The LLMs represent both state-of-the-art proprietary systems and open-source models across a range of parameter sizes and performance levels.

Two expert annotators (authors of this paper) qualitatively assess the clarity, relevance, informativeness, and verifiability of the MBCs. The identity of the system that generated each MBC is concealed from the annotators, and they are not permitted to use external sources such as search engines or LLMs to obtain additional background information about the MBCs.

\paragraph{Agreement Scores}

The Krippendorff-$\alpha$ scores for our annotations are 0.52, 0.98, 0.95, and 0.91, respectively. Furthermore, as \citet{Justus-2010:FreeMarginalKappa} argue, measures such as Fleiss-$\kappa$ assume that the distribution of assignments across score categories is fixed in advance when estimating chance agreement, an assumption that is often unrealistic in practice. We therefore also report free-marginal multi-rater kappa scores, which are 0.90, 0.93, 0.88, and 0.81 for clarity, relevance, informativeness, and verifiability, respectively.

Overall, agreement scores are high, with informativeness and verifiability being slightly lower. This is expected, as these two criteria are inherently more subjective and depend on a reader's interpretation of how an MBC affects their understanding of a news source.

\section{Analysis}
\subsection{To what extent do generated MBCs capture source credibility?}

Overall, the results of our qualitative analysis are consistent with those of our automatic evaluation. Specifically, \texttt{gpt-4o-mini} performs best at generating fully relevant MBCs, while \texttt{llama-3.3-70b-instruct} produces the most informative and verifiable outputs. In contrast, \texttt{gpt-5-mini} performs worst in terms of relevance, informativeness, and verifiability. Open-source models nevertheless remain competitive with state-of-the-art proprietary models.

\autoref{fig:qualitative_analysis} shows that all models generally produce clear and relevant MBCs. However, generating informative and verifiable content is significantly more challenging. While entirely uninformative or unverifiable MBCs (i.e., those assigned a score of 0) are rare, the scores indicate that informativeness and verifiability are harder to achieve than clarity and relevance. In particular, verifiable points must first be clear, while informative points must be both clear and relevant.

For example, consider the following excerpt from a \texttt{gpt-4o-mini}-generated MBC for \textit{Daily Surge}:

\begin{tcolorbox}[
    width=\linewidth,
    colback=blue!5,
    colframe=black!60,
    boxrule=0.5pt
]
\small
``Generally considered to have a conservative bias; aligns with right-leaning perspectives on political and social issues.''
\end{tcolorbox}

Although the model identifies a potentially right-leaning political bias, it does not specify \textit{which} right-leaning perspectives the source aligns with, making the claim difficult to verify.

\begin{figure}[t]
    \centering
    \includegraphics[width=0.95\columnwidth]{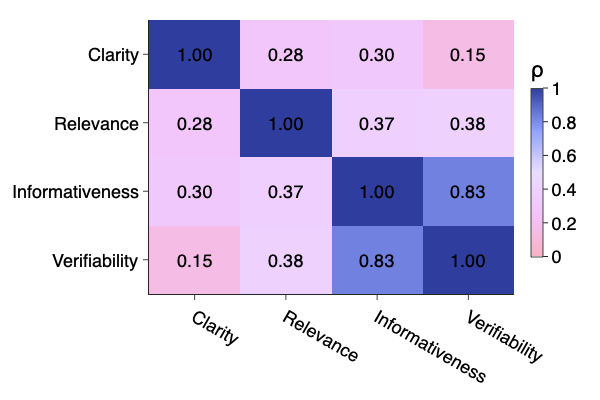}
    \caption{
        Spearman correlation scores between MBC clarity, relevance, informativeness, and verifiability.
    }
    \label{fig:qual_correlation}
\end{figure}

This observation is further supported by the results shown in \autoref{fig:qual_correlation}, which illustrate the Spearman correlation scores between the different criteria. Notably, we find that informative MBCs tend to be verifiable ($\rho = 0.83$), suggesting that providing specific evidence in an MBC may supply sufficient detail to bridge gaps in a reader's perception of the source, thereby improving verifiability scores.

\subsection{Challenges in updating MBCs}

\paragraph{Insufficient Information}

When models cannot source information about a news outlet either internally or from the knowledge store, they tend to generate weakly informative background descriptions or no useful information at all. For example, \texttt{mistral-7b-instruct} describes \textit{The Jackson Sun} in a way that is too vague and generic to meaningfully inform an assessment of source credibility:

\begin{tcolorbox}[
    width=\linewidth,
    colback=blue!5,
    colframe=black!60,
    boxrule=0.5pt
]
\small
``...like many local newspapers, [striving] to maintain a balanced approach to news reporting. However, like any media outlet, it may have editorial leanings that are not always immediately apparent.''
\end{tcolorbox}

Local newspapers appear to be particularly susceptible to this issue. Among the 34 local news outlets in our sample, 71\% received an informativeness score of at most 1, while 80\% received a verifiability score of at most 1. In many cases, the generated MBCs contained only basic background information similar to the example above. This may be due to the fact that these outlets often lack the prominence required to attract substantial external scrutiny, such as independent credibility assessments or fact-checking coverage, despite potentially being trustworthy sources for local issues. Consequently, models may be unable to provide evidence either supporting or challenging their credibility.
For instance, the same \texttt{mistral-7b-instruct}-generated MBC includes factually plausible but largely uninformative statements about the outlet's credibility:

\begin{tcolorbox}[
    width=\linewidth,
    colback=blue!5,
    colframe=black!60,
    boxrule=0.5pt
]
\small
``There is no public record of The Jackson Sun failing fact-checks by reputable fact-checking organisations.''
\end{tcolorbox}

\paragraph{Inconsistent Emphasis on Evidence}

Models differ substantially in the extent to which they interpret or evaluate the evidence they retrieve. In 8 of 27 MBCs, \texttt{gpt-5-mini} draws inferences about the relevance of evidence to source credibility, including unprompted reasoning about the source itself. By contrast, the other models rarely include content evaluating whether the available evidence sufficiently supports a conclusion (across 27 MBCs: \texttt{llama-3.3-70b-instruct} and \texttt{mistral-7b-instruct} do so in one instance each, and \texttt{gpt-4o-mini} in none). For example, for the \textit{Source New Mexico} news outlet, \texttt{gpt-5-mini} generates:
\begin{tcolorbox}[
    width=\linewidth,
    colback=blue!5,
    colframe=black!60,
    boxrule=0.5pt
]
\small
``The excerpt you provided lists many Democratic candidates and endorsers, but that listing alone does not prove an editorial endorsement or a consistent partisan tilt by the outlet.''
\end{tcolorbox}

Providing readers with information about the quality and limitations of the available evidence may help them form a more nuanced assessment of a source's credibility and is broadly consistent with professional fact-checking practices. However, such commentary may also introduce an additional interpretative layer that risks influencing readers' perceptions of the evidence. Note that this additional reasoning substantially increases the length of the MBCs, making them harder to read and understand. Specifically, median MBC word counts are 458 for \texttt{gpt-5-mini}, compared with 194 for \texttt{gpt-4o-mini}, 268 for \texttt{mistral-7b-instruct}, and 264 for \texttt{llama-3.3-70b-instruct}.

\paragraph{Information Loss}
In seven MBCs generated by \texttt{mistral-7b-instruct}, the model omits information from the beginning of the MBC during the updating process. For example, in the MBC for \textit{KJRH -- Tulsa News}, the first nine points, which describe the outlet's location, affiliation, ownership, and funding, are replaced with:

\begin{tcolorbox}[
    width=\linewidth,
    colback=blue!5,
    colframe=black!60,
    boxrule=0.5pt
]
\small
``1--9: Same as previous response.''
\end{tcolorbox}

As these early points typically contain important background information about the outlet, omitting them reduces the informativeness of the final MBC. This issue could be mitigated by explicitly instructing the model to regenerate and verify the complete updated MBC after each revision.

\paragraph{Conflicting Information}
We observe three cases (out of 108) of explicit contradictions in generated MBCs: two from \texttt{llama-3.3-70b-instruct} and one from \texttt{gpt-5-mini}. Such contradictions reduce informativeness, as they prevent readers from drawing reliable conclusions.
In two cases, contradictions arise when models update an existing MBC with newly retrieved information. For example, when describing fact-checking failures associated with \textit{WBTS -- NBC 10 Boston}, \texttt{llama-3.3-70b-instruct} initially generates \textit{``None notable found''} but later adds:

\begin{tcolorbox}[
    width=\linewidth,
    colback=blue!5,
    colframe=black!60,
    boxrule=0.5pt
]
\small
``WBTS - NBC 10 - Boston has been cited as an example of a media outlet that has distorted facts, according to a series on media bias, although specific details of the incident are not provided.''
\end{tcolorbox}

One possible explanation is the prompt design, which discourages removing previously generated content during updates. While this helps preserve earlier information, it may also allow outdated or superseded statements to remain in the MBC when later retrieved evidence introduces conflicting information. This limitation could be addressed by explicitly prompting the model to revise or remove statements that are contradicted by newly retrieved evidence.

\section{Conclusion}
We introduced \myknowledgestore{}, a publicly available knowledge store designed to support the generation of media background checks (MBCs). We described a reproducible methodology for constructing and updating the resource, and evaluated several widely used LLMs on the MBC generation task. Our experiments show that \myknowledgestore{} improves the quality of generated MBCs, while our human evaluation highlights that producing concise, informative, and verifiable MBCs remains an open challenge. 

We publicly release \myknowledgestore{} to support reproducible research on source-critical reasoning and automated fact-checking.

\section*{Limitations}

We acknowledge that search-engine rankings may introduce bias as a proxy for relevance in evidence selection, potentially affecting the quality and diversity of retrieved sources. Nevertheless, our resource provides a strong starting point that can be extended in future work to incorporate more diverse and heterogeneous data sources.

Second, although we use a blacklist to filter sources, some undesired or low-quality sources referencing MBCs or fact-checks may still be included, as no blacklist can be fully comprehensive.

Third, the system is limited to information freely available on the web and therefore cannot access restricted or proprietary sources, such as official statistical databases or paywalled archives, highlighting an opportunity for future integration of broader external sources.

\section*{Ethical Considerations}

Our resource is designed to support media background checks using publicly available information. However, despite the use of retrieval, generated outputs may still contain inaccuracies or misrepresentations of source material and should therefore not be treated as definitive judgments about the sources, especially in high-stakes settings.

Second, the reliance on web search and publicly available content introduces inherent biases, including disparities in coverage across regions, languages, and entities. Such biases may lead to uneven or incomplete representations in generated background checks. In addition, while we use filtering mechanisms such as blacklists, we do not guarantee the complete removal of low-quality or unreliable sources.

Finally, although the system operates solely on publicly accessible information, aggregating and synthesising such data may still raise dual-use concerns. The intended use of our resource \cite{schlichtkrull-etal-2023-intended} is to support research on automated fact-checking and potentially assist fact-checkers with human experts in the loop. We emphasise the need for cautious deployment and human oversight in real-world applications.
\paragraph{AI Use} We use privacy-preserving models to assist with proofreading and coding tasks (e.g., generating plots), in line with the ACL guidelines.
\bibliography{custom}
\newpage
\appendix
\section*{Appendix}
\section{Prompts}
\label{appendix:Prompts}
We give the prompts used to generate, update, and evaluate MBCs.
They are reproduced from \citet{Schlichtkrull-2024:MBCs}.

\begin{figure}[!h]
    \centering
    \fbox{\begin{minipage}{\linewidth}
    \textbf{System Message:} You are InfoHuntGPT, a world-class AI assistant used by journalists to quickly build knowledge of new sources.\\
    
    \textbf{User Message:} Build a background check for the news source [\textit{source name}].
    Write down everything you know about them, e.g. who funds them, how they make money, if they have any particular bias.
    Make an ITEMIZED LIST.
    Be brief, and if you don’t know something, just leave it out.
    If you are aware that they have failed any fact-checks, mention which.
    Begin your response with "**Background check**".
    \end{minipage}}\\
    \caption{Prompt used to generate MBCs without external evidence.
    [\textit{Source name}] stands for the name of the target news source.
    }
    \label{fig:prompt_initial_MBCs}
\end{figure}

\begin{figure}[!h]
    \centering
    \fbox{\begin{minipage}{\linewidth}
    \textbf{Assistant Message:} [\textit{Previous MBC}]\\
    
    \textbf{User Message:} Google search has revealed some new information:
    [\textit{New information}]\\
    Update your background check for [\textit{source name}] using the new information.
    Do NOT delete any information, but make ADDITIONS where necessary, using the new information.
    Most likely, you will just need to add an extra item to the itemized list you previously created.
    Make minimal edits, and only incorporate what is relevant.
    Begin your response with "**Background check**"
    \end{minipage}}
    \caption{Prompt used to update MBCs when provided with external evidence.
    [\textit{Source name}] stands for the name of the target news source. [\textit{New information}] represents the retrieved information to be incorporated.
    }
    \label{fig:prompt_update_MBCs}
\end{figure}

\begin{figure}[!h]
    \centering
    \fbox{\begin{minipage}{\linewidth}
    \textbf{System Message:} You are FactCheckGPT, a world-class tool used by journalists to discover
    problems in their writings. Users give you text, and check whether facts are true given the text.
    You ALWAYS answer either TRUE, FALSE, or NOT ENOUGH EVIDENCE.\\
    
    \textbf{User Message:} You will be given a snippet written as part of a source criticism exercise, and a
    claim. Your task is to determine whether the claim is true based ONLY on the text. Do NOT use any other knowledge source.\\
    
    The claim is: [\textit{question}]. \\
    The text follows below: [\textit{text}].\\
    
    [\textit{question}]? Thinking step by step, answer either TRUE, FALSE, or NOT ENOUGH EVIDENCE,
    capitalizing all letters. Explain your reasoning FIRST, and after that output either TRUE, FALSE,
    or NOT ENOUGH EVIDENCE.
    \end{minipage}}
    \caption{Prompt used to determine whether an atomic fact is entailed or contradicted by a
    generated MBC. [\textit{question}] is replaced with the atomic fact template, and [\textit{text}] is replaced by the generated MBC.
    }
    \label{fig:prompt_atomic_facts}
\end{figure}

\section{Search Queries for Evidence Retrieval}
\label{appendix:SearchQueries}
\autoref{table:BaselineSearchQueries} gives the search queries used to retrieve evidence from \myknowledgestore{} when generating MBCs with information retrieval.
\begin{table*}[!h]
\centering
\begin{tabular}{lll}\toprule
    \#  &    Query &     Question \\ \midrule
    1   &    ``{source name}'' ownership &  Who owns ``{source name}''?\\
    2   &    ``{source name}'' funding &  How is ``{source name}'' funded?\\
    3   &    ``{source name}'' about &  What is ``{source name}''?\\
    4   &    ``{source name}'' political leaning &  What is the political leaning of ``{source name}''?\\
    5   &    ``{source name}'' fact-check &  Has ``{source name}'' failed any fact-checks?\\
    6   &    ``{source name}'' retracted article &  Has ``{source name}'' retracted any articles?\\ \bottomrule
\end{tabular}
\caption{Queries used to retrieve evidence from \myknowledgestore{}, with corresponding questions used to extract answers with DeBERTa. Reproduced from \citet{Schlichtkrull-2024:MBCs}.}
\label{table:BaselineSearchQueries}
\end{table*}

\section{Atomic Fact Templates}
\label{appendix:Atomic Facts}
\autoref{table:atomic_templates} gives the 42 atomic facts used to evaluate the fact recall and error rate metrics.

\begin{table*}[!h]
\centering
\begin{tabular}{@{}ll@{}}
\toprule
Type                    & Template \\
\midrule
founded\_source         & \_\_ founded "source name" \\
host                    & \_\_ hosts "source name" \\
founder\_of\_source     & \_\_ is the founder of "source name" \\
ceo\_of\_source         & \_\_ is the ceo of "source name" \\
leads\_source           & \_\_ leads "source name" \\
owns\_source            & \_\_ owns "source name" \\
publishes\_source       & \_\_ publishes "source name" \\
owner\_of\_source       & \_\_ is the owner of "source name" \\
bought\_source          & \_\_ bought "source name" \\
acquired\_source        & \_\_ acquired "source name" \\
source\_funded\_through & "source name" is funded through \_\_ \\
remit                   & The remit of "source name" is to \_\_ \\
usual\_audience         & The usual audience of "source name" is \_\_ \\
awards                  & "source name" was awarded \_\_ \\
source\_tends\_to       & "source name" tends to the \_\_ \\
source\_other\_comments & Other sources have commented on "source name", stating that it tends to be \_\_ \\
factuality              & The factuality of "source name" is reported to be \_\_ \\
article\_failed\_check  & "source name" failed a fact-check for an article titled \_\_ \\
article\_retraction     & "source name" printed a retraction after failing a fact-check for an article titled \_\ \\
topic\_not\_covered     & "source name" chose not to cover \_\_ \\
is\_original\_content   & "source name" provides original content written by staff journalists \\
uses\_advertising       & "source name" relies on advertising for revenue \\
uses\_subscriptions     & "source name" relies on subscriptions for revenue \\
uses\_donations         & "source name" relies on donations for revenue \\
donations\_from         & "source name" has received donations from \_\_ \\
aggregation\_source     & When aggregating stories, "source name" relies on information from \_\_ \\
has\_endorsed           & "source name" has endorsed \_\_ \\
editorial\_bias         & "source name" has an editorial bias towards \_\_ \\
editor                  & \_is the editor of "source name" \\
fined\_amount\_reason   & "source name" was fined \_\_ for \_\_ \\
biased\_article         & "source name" printed a biased article titled \_\_ \\
misleading\_article     & "source name" printed a factually misleading article titled \_\_ \\
apologized\_for         & "source name" had to apologize for \_\_ \\
damages\_to\_reason     & "source name" paid damages to \_\_ after \_\_ \\
pretends\_to\_be        & "source name" pretends to be \_\_ \\
unknown\_who            & It is unknown who \_\_ \\
headquarters\_location  & "source name"'s headquarter is located in \_\_ \\
gov\_funded             & "source name" is funded by the \_\_ government \\
source\_is              & "source name" is a \_\_ \\
uses\_peer\_review      & "source name" uses a peer review process \\
uses\_int\_checks       & "source name" uses an internal fact-checking process \\
covers\_topics          & "source name" covers the following topics: \_\_ \\ 
\bottomrule
\end{tabular}
\caption{Templates used to extract atomic facts from MBCs, by filling in the blanks. Reproduced from \citet{Schlichtkrull-2024:MBCs}.}
\label{table:atomic_templates}
\end{table*}

\section{Qualitative Annotation Guidelines}
\label{appendix:Annotation Criteria}
The four criteria are defined as follows.
\paragraph{Clarity:} Background checks should be clear and understandable to an average layperson.
\begin{itemize}[noitemsep,nolistsep]
    \item \textbf{0 -- Poor clarity:} The MBC's language or format severely hinders comprehension. Most points are unclear in meaning. Examples include ungrammatical text that impedes understanding, omitted information necessary to interpret later points, or contradictory statements regarding source credibility.
    \item \textbf{1 -- Limited clarity:} Most points are understandable, but language or formatting occasionally hinders comprehension.
    \item \textbf{2 -- High clarity:} All points are clear and understandable.
\end{itemize}

\paragraph{Relevance:} All facts mentioned in the check should relate to the target source or closely connected entities and topics.
\begin{itemize}[noitemsep,nolistsep]
    \item \textbf{0 -- Off-topic:} The MBC as a whole is unrelated to the target source.
    \item \textbf{1 -- Partially relevant:} Some, but not all, points refer to unrelated entities or ideas and are off-topic.
    \item \textbf{2 -- Broadly relevant:} Some points refer to entities or ideas not directly related to the source but still relevant to the broader topic.
    \item \textbf{3 -- Fully relevant:} All points are directly related to the target source and remain on topic.
\end{itemize}

\paragraph{Informativeness:} Facts about the target source should influence the reader's understanding of the source's general credibility and provide context relevant to evaluating claims.
Because an MBC evaluates a source overall rather than a specific claim, highly specific facts are less useful.
\begin{itemize}[noitemsep,nolistsep]
    \item \textbf{0 -- Uninformative or contradictory:} Points do not indicate source credibility, or conflicting information prevents conclusions from being drawn confidently.
    \item \textbf{1 -- Weakly informative:} Points describe characteristics only weakly related to source credibility (e.g.\ general background information), with unclear implications for credibility.
    \item \textbf{2 -- Moderately informative:} At least one point describes source characteristics relevant to credibility, although their implications are not fully explicit or apply only to limited claim types. No contradictory points are present.
    \item \textbf{3 -- Highly informative:} At least one point directly indicates the credibility and political bias (or absence of bias) of most claims from the source and explains why. No contradictory points are present.
\end{itemize}

\paragraph{Verifiability:} Sufficient evidence should be provided to support factual claims, or enough detail included that a layperson could reasonably locate the evidence independently.
Annotators should not conduct additional research and should base judgements solely on the ease with which facts could plausibly be verified.
\begin{itemize}[noitemsep,nolistsep]
    \item \textbf{0 -- Unverifiable:} Points are entirely unverifiable, such as broad subjective assertions.
    \item \textbf{1 -- Limited verifiability:} Points provide sufficient detail for further investigation, but omit important contextual information needed to assess them independently.
    \item \textbf{2 -- Moderately verifiable:} Points are sufficiently detailed to appear credible and be verifiable by a layperson, although verification would require some effort.
    \item \textbf{3 -- Easily verifiable:} Specific evidence is provided that a layperson could quickly and easily verify.
\end{itemize}

\section{Implementation Details}
All code was implemented in Google Colab with an NVIDIA T4 GPU.

\subsection{Online Evidence Retrieval}
URLs for evidence documents were retrieved via the Google Custom Search JSON API\footnote{\url{https://developers.google.com/custom-search/v1/overview}}.
Text was web-scraped via the \texttt{trafilatura} package, using the default config file provided.
Up to two attempts to web-scrape each URL were performed.

\subsection{MBC Generation}
We access the LLMs for generating and updating MBC text using the OpenRouter\footnote{\url{https://openrouter.ai}} API, via the OpenAI SDK.
The DeBERTa model in the +IR setting is implemented via a HuggingFace pipeline with the `question-answering' task\footnote{\url{https://huggingface.co/deepset/deberta-v3-large-squad2}}.
Parameters for all models were left as their defaults.

\begin{table}[ht]
\begin{tabular}{@{}llc@{}}
\toprule
\textbf{\texttt{Hyperparameter}} & \textbf{\texttt{Range}} & \textbf{\texttt{Default Value}} \\ \midrule
\textbf{\texttt{Temperature}} & {[}0.0, 2.0{]} & 1.0 \\
\textbf{\texttt{Top P}} & {[}0.0, 1.0{]} & 1.0 \\
\textbf{\texttt{Top K}} & $\geq$ 0 & 0 \\
\textbf{\texttt{Frequency Penalty}} & {[}-2.0, 2.0{]} & 0.0 \\
\textbf{\texttt{Presence Penalty}} & {[}-2.0, 2.0{]} & 0.0 \\
\textbf{\texttt{Repetition Penalty}} & {[}0.0, 2.0{]} & 0.0 \\
\textbf{\texttt{Min P}} & {[}0.0, 1.0{]} & 0.0 \\
\textbf{\texttt{Top A}} & {[}0.0, 1.0{]} & 0.0 \\
\textbf{\texttt{Verbosity}} &  & medium \\ \bottomrule
\end{tabular}
\caption{Default values used when calling LLMs via OpenRouter to generate MBC text.}
\label{table:LLM_hyperparameters}
\end{table}

We use the \texttt{BM25Okapi} algorithm from the \texttt{rank\_bm25} package to retrieve evidence from the knowledge store.
We perform word-level and sentence-level chunking using the \texttt{word\_tokenize} function from NLTK and the \texttt{Sentencizer} from \texttt{spaCy} respectively.

\subsection{MBC Evaluation}
To calculate ROUGE and METEOR scores, we use the `rougeL' algorithm from the \texttt{rouge\_score} package with `\texttt{use\_stemmer} = True', and the `meteor\_score' algorithm from \texttt{NLTK}.

For atomic fact-based metrics, we access the \texttt{GPT-3.5-Turbo} model through OpenRouter and the OpenAI SDK.

\end{document}